\documentclass[runningheads]{llncs}
\usepackage[T1]{fontenc}
\usepackage{xcolor}
\usepackage{graphicx}

\begin{document}

\title{Machine Apophenia: The Kaleidoscopic Generation of Architectural Images}


\author{Tikhonov Alexey\inst{1}\orcidID{0000-0001-6972-5171} \and Dmitry Sinyavin\inst{2}}
\authorrunning{Tikhonov A. and Sinyavin D.}

\institute{Inworld.AI, Germany \\
\email{altsoph@gmail.com}
\and Independent Researcher
\email{s0me0ne@s0me0ne.com}
} 

% \author{anonymized}
% \authorrunning{anonymized}

\maketitle              % typeset the header of the contribution
\begin{abstract}
This study investigates the application of generative artificial intelligence in architectural design. We present a novel methodology that combines multiple neural networks to create an unsupervised and unmoderated stream of unique architectural images. Our approach is grounded in the conceptual framework called machine apophenia. We hypothesize that neural networks, trained on diverse human-generated data, internalize aesthetic preferences and tend to produce coherent designs even from random inputs. The methodology involves an iterative process of image generation, description, and refinement, resulting in captioned architectural postcards automatically shared on several social media platforms. Evaluation and ablation studies show the improvement both in technical and aesthetic metrics of resulting images on each step.

\end{abstract}
\section{Introduction}

The "Freaking Architecture" project%\footnote{Link omitted for anonymity} 
\footnote{\url{https://altsoph.github.io/freaking-architecture/}} 
represents a novel approach to the application of generative AI in creative fields, particularly architecture. This study explores the intersection of artificial intelligence and creative design, utilizing an ensemble of neural networks to generate an unsupervised and unmoderated stream of unique architectural images.

The project employs random keyword combinations to create captioned architectural postcards. Beginning with seed keyphrases derived from 50 selected architectural images, the process expands upon these initial inputs using various AI models, including CLIP\cite{radford2021learning}, BLIP2\cite{li2023blip2}, StableDiffusionV1\cite{rombach2021highresolution} and SDXL\cite{stablediffusion2022}. The resulting images, shared across social media platforms, demonstrate the potential of AI in creative visual content generation.

% [This research contributes to the broader discourse on AI's role in creative processes, exploring how machine learning algorithms can interpret and expand upon human-generated concepts in architecture.]
% \section{Background}

The application of neural networks in generative tasks spans various creative domains, including literature\cite{fi12110182}, typography\cite{liao2023callipaint,pippi2023handwritten,fi12110182}, visual arts\cite{shahriar2021gan,romero2021neural}, and music\cite{nierhaus2009algorithmic,complexis21}. These AI-driven approaches to content creation raise important questions about the nature of creativity and the potential for machines to produce aesthetically pleasing outputs.

A critical challenge in AI-generated content is the evaluation of results, particularly when assessing aesthetic qualities. While some pre-trained models exist for evaluating technical and aesthetic aspects of images\cite{idealods2018imagequalityassessment}, the subjective nature of aesthetics complicates this process\cite{10.1145/3539618.3592000}.

\section{Framework}

This study proposes a novel conceptual approach that combines two key ideas: the kaleidoscope effect and machine apophenia. These concepts form the foundation for our methodology of unsupervised and unmoderated generation of high-quality architectural content.

The \textbf{kaleidoscope effect} draws inspiration from complex systems theory, particularly the emergence of ordered patterns from simple rules in cellular automata\cite{wolfram2002new}. This concept suggests that intricate and aesthetically pleasing patterns can arise from basic mechanistic recombinations, analogous to the way the I Ching derives complex interpretations from simple coin tosses\cite{wilhelm1967iching}. In our context, the kaleidoscope effect represents the potential for AI systems to generate diverse and coherent architectural designs through iterative processes applied to random initial inputs.

It is known that different training datasets could influence biases of AI models\cite{pessach2020mitigating,bai2023comprehensive}. Building on the analogy with a psychological concept of apophenia - the human tendency to perceive meaningful patterns in random data - we propose the \textbf{machine apophenia hypothesis}. This hypothesis suggests that neural networks, when trained on diverse human-generated data, internalize human aesthetic preferences and cultural norms. Consequently, even when starting from random noise or arbitrary inputs, the network is inclined to adjust its output towards something that aligns with human aesthetic sensibilities, demonstrating some form of "aesthetic bias." 

The combination of these two concepts - the kaleidoscope effect and machine apophenia - provides a foundation for understanding how AI systems can generate aesthetically pleasing and coherent architectural designs from random or minimal inputs. This framework suggests that the iterative application of AI models combined with learned aesthetic preferences can result in outputs that are both novel and aligned with human aesthetic sensibilities.
\begin{figure}
\includegraphics[width=\textwidth]{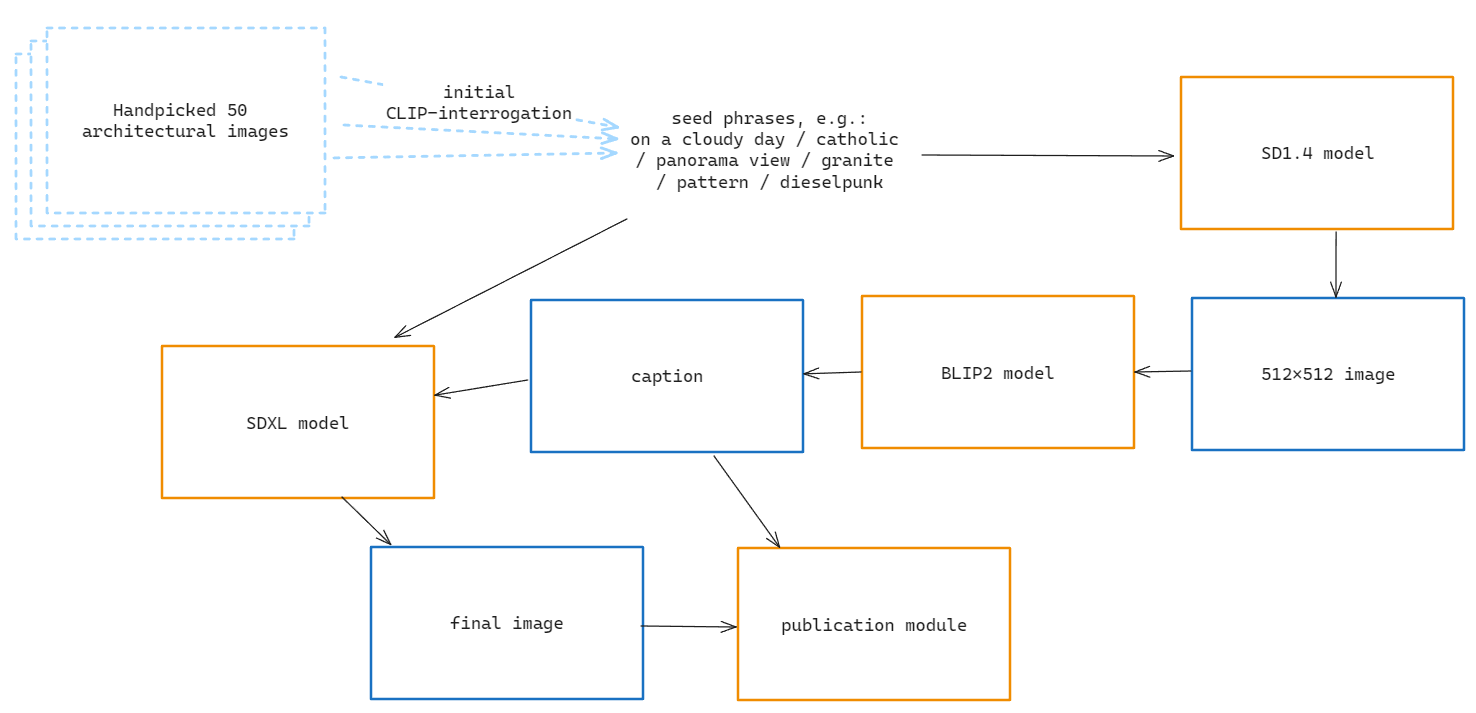}
\caption{An overall flowchart of the process.} \label{fig1}
\end{figure}

\section{Approach}

% (see Fig.~\ref{fig1}).

Building on our framework, we employ an multi-step process where random inputs are repeatedly processed through various neural networks, resulting in increasingly refined and aesthetically organized outputs. This method bears distant similarities to the refinement process in modern diffusion models, where random noise gradually converges toward meaningful images.
\\ \\ 
Our methodology can be broken down into the following steps:

% \begin{itemize}
\begin{enumerate}
\item Initial Seed Generation:\begin{itemize}
   \item We start with a curated dataset of 50 diverse architectural images.
   \item We apply the CLIP-interrogation technique\cite{pharmapsychotic2021clipinterrogator} to generate initial seed keyphrases from these images, capturing essential features and styles.
   \item We then use the GPT-3 model\cite{brown2020language} to generate expanded sets of related architectural terms and concepts, ending with 408 keyphrases. This expansion process creates a diverse pool of architectural vocabulary, enriching the subsequent generation steps.
\end{itemize}

\item Image Generation:\begin{itemize}
   \item First, we randomly pick several keyphrases from our seed pool and compile the initial query from them. % (see Table~\ref{tab1} for examples.)
   \item Then, we use the Stable Diffusion v1.4 model without any fine-tuning to generate initial 512x512 images based on the generated queries.
   % - This step leverages the model's ability to interpret textual descriptions and translate them into visual representations.
\end{itemize}

\item Query Refinement:\begin{itemize}
   \item We utilize the BLIP2 model to generate textual descriptions of the AI-created images. Next, we combine these new descriptions with the initial keyword-based queries to get the refined queries.
   \item This step creates a feedback loop, allowing the system to interpret and describe its own creations, potentially introducing new concepts or refining existing ones.
\end{itemize}
\item Image Refinement:\begin{itemize}
   \item The refined queries from the previous step are fed back into SDXL (Stable Diffusion XL) to create refined versions of the architectural images, sized as 1024x1024.
   \item We use no quality control or filtering at all, except the safety filter built-in in the SDXL model.
   % \item Note: at this step. one can try to combine the textual input with image input, utilizing initial images. See the ablation studies section for the details.
\end{itemize}
\begin{figure}
\begin{center}
\includegraphics[width=0.82\textwidth]{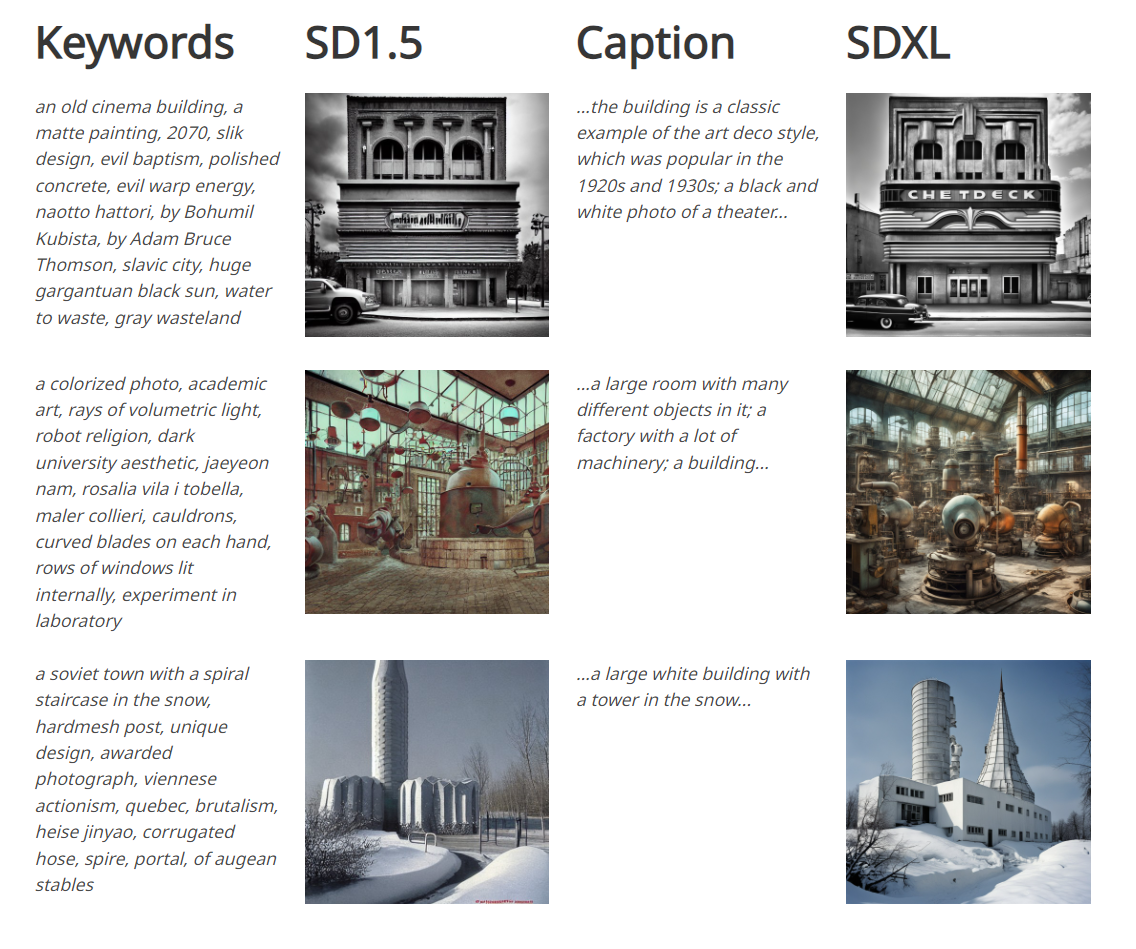}
\end{center}
\caption{Comparative analysis of refinement stages results.} \label{stages}
\end{figure}
\item Publishing:\begin{itemize}
   \item The final images are paired with their corresponding textual descriptions to create "architectural postcards."
   \item A template-based approach is used to ensure consistent formatting and presentation for each of the used media (namely: 
%   Telegram\footnote{Link omitted for anonymity}, mastodon\footnote{Link omitted for anonymity}, tumblr\footnote{Link omitted for anonymity}, bluesky\footnote{Link omitted for anonymity}
Telegram\footnote{\url{https://t.me/s/freakingarchitecture}}, mastodon\footnote{\url{https://botsin.space/@freakingarchitecture}}, tumblr\footnote{\url{https://www.tumblr.com/freaking-architecture}}, bluesky\footnote{\url{https://bsky.app/profile/freakingarch.bsky.social}}) (see Appendix A.).
   % \item 
   We apply a stochastic publication schedule (known to have positive effects on user engagement\cite{goetgeluk2021effectiveness},) publishing another item with the probability of 1/80 each minute from 10:00 till 18:59 CET. That gives us a mean of 6.75 images published daily.
\end{itemize}
\end{enumerate}

A comparative analysis of the result of each of the refinement stages is shown in Fig.~\ref{stages}).

% \begin{table}
% \caption{Examples of random queries generated from the pool of keywords.}\label{tab1}
% \small{
% \begin{tabular}{l} %|l|l|l|}
% \hline
% an abstract sculpture with black vertical slatted timber, stunningly ominous,\\
% panorama view, unique architecture, the mines of moria, catholic, disconnected\\ shapes, by Yuri Ivanovich Pimenov, by Taravat Jalali Farahani, by Ben Enwonwu,\\ houses and roads, russian city of the future, soviet military, chrome skeksis\\
% \hline
% palace, 16mm grain, classified photo, national geograph, cd cover artwork,\\ vril, dream wave aesthetic, well-appointed space, by Oswaldo Guayasamín,\\ maia sandu, by Ben Enwonwu, experiment in laboratory, houses in the shape\\ of mushrooms, pattern, holy cross\\
% \hline
% a concrete structure in the winter, awarded photograph, winner of design\\ award, archillect, unreal rendered, neo-romanticism, evil baptism, polished\\ concrete, phong yintion j - jiang geping, by Matthias Weischer, rankin, \\spirals, granite, spire
% % Heading level %&  Example & Font size and style\\
% % \hline
% % Title (centered) &  {\Large\bfseries Lecture Notes} & 14 point, bold\\
% % 1st-level heading &  {\large\bfseries 1 Introduction} & 12 point, bold\\
% % 2nd-level heading & {\bfseries 2.1 Printing Area} & 10 point, bold\\
% % 3rd-level heading & {\bfseries Run-in Heading in Bold.} Text follows & 10 point, bold\\
% % 4th-level heading & {\itshape Lowest Level Heading.} Text follows & 10 point, italic\\
% % \hline
% \end{tabular}
% }
% \end{table}

\section{Evaluation}

The study utilizes both technical and aesthetic metrics to evaluate the generated images. To comprehensively assess the quality and relevance of the generated architectural images, we use Image Quality Assessment metrics developed in \cite{idealods2018imagequalityassessment}. Specifically, we apply pre-trained "aesthetic" and "technical" models, and each returns a score from 1 to 10. The first model aims to address the aesthetic aspects of the image, and the second tries to evaluate the "clearness" of the picture (in terms of visual artifacts).

We used both to evaluate our approach and to show how each of its steps improves both scores. To isolate the contributions of various system components, we conducted a series of ablation studies comparing different model configurations:
\begin{enumerate}
    \item \textbf{Our schema} -- described in the Approach section above.
    \item \textbf{SD14 only} -- an ablated setup that outputs the initial image generated by the StableDiffusion v1.4 model.
    \item \textbf{SDXL from keyphrases} -- an ablated setup that uses initial keyphrases to query SDXL directly, without the query refinement stage.
    \item \textbf{SDXL from img} -- an ablated setup that uses the initial image to query SDXL.
\end{enumerate}

We used 1000 images produced by each pipeline version to evaluate average scores. The results can be found in Table~\ref{table:metrics}.

\begin{table}[h!]
    \centering
    \begin{tabular}{|c|c|c|}
        \hline
        \textbf{schema} & \textbf{aesthetic score} & \textbf{technical score} \\
        \hline
        our schema & \textbf{5.91} & \textbf{5.20} \\
        \hline
        % sdxl & 5.8549 & 5.1912 \\
        SDXL~from~keyphrases  & 5.88 & 5.18 \\
        SDXL~from~img & 5.69 & 5.12 \\
        SD14~only & 5.57 & 5.01 \\
        \hline
    \end{tabular}
    \caption{Quality Metrics}
    \label{table:metrics}
\end{table}

The key observation from this evaluation is that each system component contributes significantly to the final result. Each step in our multi-step process contributes to the overall enhancement of the generated images. The iterative refinement stages, from initial seed generation to final image refinement, result in higher aesthetic and technical scores compared to more straightforward, single-step approaches.
% The evaluation metrics show that our approach not only enhances the aesthetic appeal of the images but also maintains a high level of technical quality.
% The multi-step process is crucial for generating diverse, refined, and architecturally interesting designs.

\section{Observational Study}

To understand the impact of different keyphrases on user engagement, we conducted a factor analysis based on one year of data collected from emoji reactions on our Telegram channel. This analysis aimed to identify which keyphrases significantly influenced user engagement, measured through the conversion rate of emoji feedback. The conversion rate was calculated as the percentage of images that received emoji reactions out of the total number of images containing a specific keyphrase. This metric served as a proxy for user engagement.

The results of this analysis are summarized in Table~\ref{table:engagement_keyphrases}. The table categorizes keyphrases into two groups: significantly engaging keyphrases and less engaging keyphrases, based on their conversion rates.

\begin{table}[h!]
    \centering
    \begin{tabular}{|c|c|}
        \hline
        \textbf{Keyphrase} & \textbf{Conversion Rate (\%)} \\
        \hline
        \textit{average}  & 23 \\
        \hline
        \multicolumn{2}{|c|}{\textbf{More engaging keyphrases}} \\
        \hline
        three strange objects & 63 \\
        a city at night & 49 \\
        a creepy secret temple & 43 \\
        dirty laboratory & 38 \\
        a massive cathedral in a forest & 38 \\
        thundercats & 36 \\
        a matte painting & 36 \\
        alena aenami & 35 \\
        slum & 35 \\
        a detailed matte painting & 34 \\
        made of rubber & 34 \\
        an abstract sculpture & 34 \\
        endless lake & 34 \\
        \hline
        \multicolumn{2}{|c|}{\textbf{Less engaging keyphrases}} \\
        \hline
        high-definition photograph & 14 \\
        neoclassical & 13 \\
        savannah & 13 \\
        vintage showcase of the 60s & 13 \\
        granite & 13 \\
        a building & 12 \\
        black vertical slatted timber & 11 \\
        photo taken from above & 11 \\
        an old cinema building & 11 \\
        a very tall building & 11 \\
        khedival opera house & 10 \\
        high tech concrete bench cube & 6 \\
        \hline
    \end{tabular}
    \caption{Keyphrases and their conversion rates}
    \label{table:engagement_keyphrases}
\end{table}

The analysis reveals that certain keyphrases significantly enhance user engagement. For instance, keyphrases like "three strange objects" (63\%), "a city at night" (49\%), and "a creepy secret temple" (43\%) have high conversion rates, indicating strong user interest. These keyphrases likely evoke curiosity and emotional responses, leading to higher engagement.

Conversely, keyphrases such as "high tech concrete bench cube" (6\%), "khedival opera house" (10\%), and "a very tall building" (11\%) show lower conversion rates. These phrases may be perceived as less intriguing or visually stimulating, resulting in reduced user interaction.

\begin{figure}
\begin{center}
\includegraphics[width=0.8\textwidth]{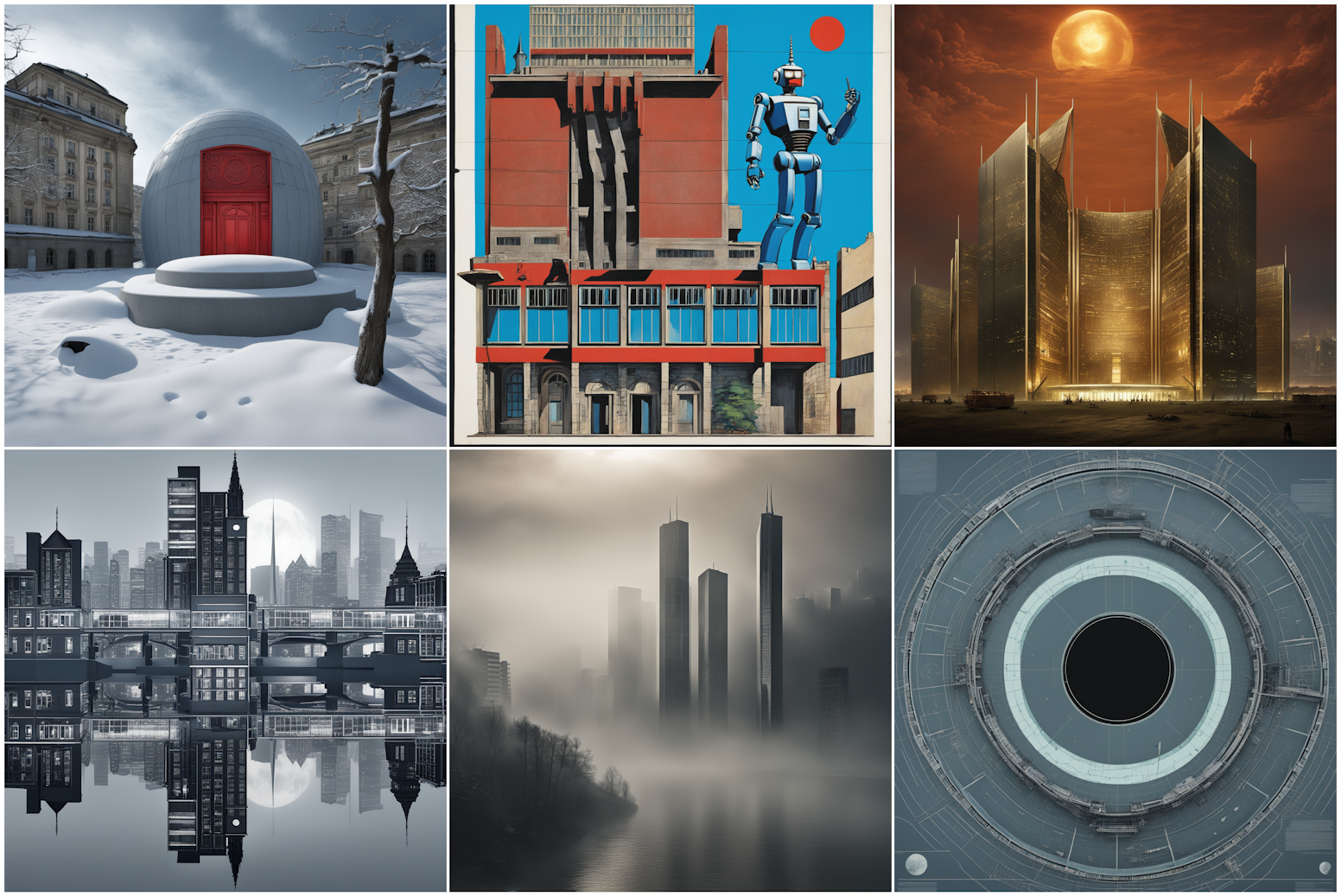}
\end{center}
\caption{Examples of images corresponding to queries with high concentrations of high-engaging keyphrases.} \label{best}
\end{figure}

\begin{figure}
\begin{center}
\includegraphics[width=0.8\textwidth]{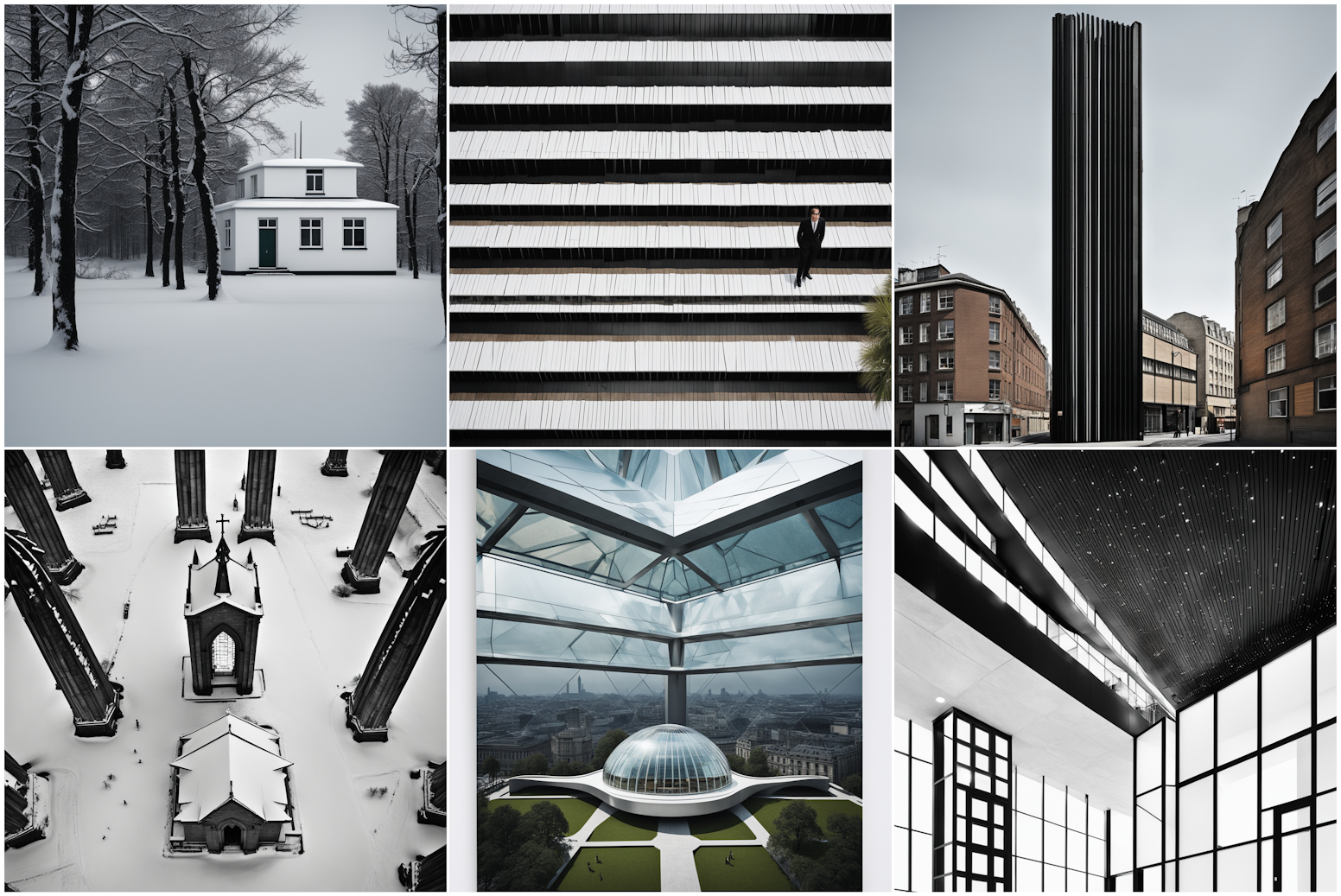}
\end{center}
\caption{Examples of images corresponding to queries with high concentrations of low-engaging keyphrases.} \label{worst}
\end{figure}

Figures \ref{best} and \ref{worst} illustrate images corresponding to queries with high concentrations of high and low engaging keyphrases, respectively. These visual representations further highlight the differences in user engagement based on the keyphrases used.

Several speculative observations can be made from this analysis. Keyphrases that evoke a sense of mystery, complexity, or artistic value tend to engage users more effectively. In contrast, keyphrases that are more technical or mundane appear to attract less attention. This insight can guide future content generation strategies, emphasizing the use of more engaging keyphrases to enhance user interaction and satisfaction.

\section{Discussion}

The findings of this study highlight the potential of generative AI in the field of architectural design, mainly through the concepts of the kaleidoscope effect and machine apophenia. By leveraging these concepts, we have demonstrated that AI systems can produce aesthetically pleasing and coherent architectural designs from random or minimal inputs. This approach not only broadens the scope of AI applications in creative fields but also raises important questions about the nature of creativity and the role of human input in AI-generated content.

One of the key insights from our study is the iterative refinement process, which significantly enhances the quality of the generated images. The multi-step methodology, involving initial seed generation, image generation, query refinement, and image refinement, ensures that the final outputs are both technically sound and aesthetically appealing. 

The evaluation metrics used in this study, particularly the aesthetic and technical scores, provide a comprehensive assessment of the generated images. The ablation studies further underscore the importance of each component in our system, demonstrating that the combination of various neural networks and iterative refinement leads to superior results. The comparative analysis of different model configurations reveals that our proposed schema consistently outperforms other setups, highlighting the effectiveness of our approach.

The observational study on user engagement offers valuable insights into the impact of different keyphrases on user interaction. The analysis indicates that keyphrases evoking curiosity, mystery, and artistic value tend to engage users more effectively. This finding can inform future content generation strategies, emphasizing the use of more engaging keyphrases to enhance user interaction and satisfaction.

However, the study also acknowledges the challenges associated with evaluating AI-generated content, particularly in terms of aesthetic qualities. The subjective nature of aesthetics complicates the evaluation process, and while pre-trained models provide a helpful starting point, they may not fully capture the nuances of human aesthetic preferences. Future research could explore more sophisticated evaluation techniques, incorporating user feedback and other qualitative measures to better assess the aesthetic value of AI-generated content.

\begin{credits}
\subsubsection{\ackname}
We thank Vadim Nikulin for his help with Telegram users feedback collection.
\end{credits}

% \subsubsection{\discintname}
% It is now necessary to declare any competing interests or to specifically
% state that the authors have no competing interests. Please place the
% statement with a bold run-in heading in small font size beneath the
% (optional) acknowledgments\footnote{If EquinOCS, our proceedings submission
% system, is used, then the disclaimer can be provided directly in the system.},
% for example: The authors have no competing interests to declare that are
% relevant to the content of this article. Or: Author A has received research
% grants from Company W. Author B has received a speaker honorarium from
% Company X and owns stock in Company Y. Author C is a member of committee Z.
% \end{credits}

\bibliographystyle{splncs04}

\bibliography{custom}

\begin{thebibliography}{10}
\providecommand{\url}[1]{\texttt{#1}}
\providecommand{\urlprefix}{URL }
\providecommand{\doi}[1]{https://doi.org/#1}

\bibitem{fi12110182}
Agafonova, Y., Tikhonov, A., Yamshchikov, I.P.: Paranoid transformer: Reading narrative of madness as computational approach to creativity. Future Internet  \textbf{12}(11) (2020). \doi{10.3390/fi12110182}, \url{https://www.mdpi.com/1999-5903/12/11/182}

\bibitem{stablediffusion2022}
AI, S.: Stable diffusion v2. \url{https://github.com/Stability-AI/stablediffusion} (2022), accessed: 2024-06-22

\bibitem{bai2023comprehensive}
Bai, M., Guo, Y., Nie, L., Cheng, H., Cheng, Z., Kankanhalli, M., Bimbo, A.: A comprehensive review of bias in deep learning models: Methods, impacts, and future directions. Archives of Computational Methods in Engineering  \textbf{30},  3544--3556 (2023)

\bibitem{brown2020language}
Brown, T.B., Mann, B., Ryder, N., Subbiah, M., Kaplan, J.D., Dhariwal, P., Neelakantan, A., Shyam, P., Sastry, G., Askell, A., et~al.: Language models are few-shot learners. arXiv preprint arXiv:2005.14165  (2020)

\bibitem{goetgeluk2021effectiveness}
Goetgeluk, S., Broers, G., Van~der Eecken, H., Wouters, E., De~Vriendt, P.: Effectiveness of an intervention for reducing sitting time and improving health in office workers: three arm cluster randomised controlled trial. BMJ  \textbf{372} (2021)

\bibitem{idealods2018imagequalityassessment}
Lennan, C., Nguyen, H., Tran, D.: Image quality assessment. \url{https://github.com/idealo/image-quality-assessment} (2018)

\bibitem{li2023blip2}
Li, J., Li, D., Xiong, C., Hoi, S.: Blip-2: Bootstrapping language-image pre-training with frozen image encoders and large language models. arXiv preprint arXiv:2301.12597  (2023)

\bibitem{liao2023callipaint}
Liao, Q., Wang, Z., Abdul-Mageed, M., Xia, G.: Callipaint: Chinese calligraphy inpainting with diffusion model (2023)

\bibitem{nierhaus2009algorithmic}
Nierhaus, G.: Algorithmic Composition: Paradigms of Automated Music Generation. Springer Vienna (2009). \doi{10.1007/978-3-211-75540-2}

\bibitem{10.1145/3539618.3592000}
Pavlichenko, N., Ustalov, D.: Best prompts for text-to-image models and how to find them. In: Proceedings of the 46th International ACM SIGIR Conference on Research and Development in Information Retrieval. p. 2067–2071. SIGIR '23, Association for Computing Machinery, New York, NY, USA (2023). \doi{10.1145/3539618.3592000}, \url{https://doi.org/10.1145/3539618.3592000}

\bibitem{pessach2020mitigating}
Pessach, D., Shmueli, E.: Through a fair looking-glass: Mitigating bias in image datasets. In: Proceedings of the IEEE/CVF Conference on Computer Vision and Pattern Recognition. pp. 8227--8236 (2020)

\bibitem{pharmapsychotic2021clipinterrogator}
Pharmapsychotic: Clip interrogator: Image to prompt with clip. \url{https://github.com/pharmapsychotic/clip-interrogator} (2021), gitHub repository

\bibitem{pippi2023handwritten}
Pippi, V., Cascianelli, S., Cucchiara, R.: Handwritten text generation from visual archetypes (2023)

\bibitem{radford2021learning}
Radford, A., Kim, J.W., Hallacy, C., Ramesh, A., Goh, G., Agarwal, S., Sastry, G., Askell, A., Mishkin, P., Clark, J., et~al.: Learning transferable visual models from natural language supervision. arXiv preprint arXiv:2103.00020  (2021)

\bibitem{rombach2021highresolution}
Rombach, R., Blattmann, A., Lorenz, D., Esser, P., Ommer, B.: High-resolution image synthesis with latent diffusion models (2021)

\bibitem{romero2021neural}
Romero, J., Machado, P.: Neural networks in art, sound and design. Neural Computing and Applications  \textbf{33},  1--20 (2021)

\bibitem{shahriar2021gan}
Shahriar, S.: Gan computers generate arts? a survey on visual arts, music, and literary text generation using generative adversarial networks. Pattern Recognition Letters  (2021)

\bibitem{complexis21}
Tikhonov., A., Yamshchikov., I.P.: Artificial neural networks jamming on the beat. In: Proceedings of the 6th International Conference on Complexity, Future Information Systems and Risk - COMPLEXIS. pp. 37--44. INSTICC, SciTePress (2021). \doi{10.5220/0010461200370044}

\bibitem{wilhelm1967iching}
Wilhelm, R.: The I Ching or Book of Changes. Princeton University Press, Princeton, NJ (1967), translated by Richard Wilhelm, rendered into English by Cary F. Baynes

\bibitem{wolfram2002new}
Wolfram, S.: A New Kind of Science. Wolfram Media, Champaign, IL (2002)

\end{thebibliography}





% LNCS Homepage, \url{http://www.springer.com/lncs}, last accessed 2023/10/25
% \end{thebibliography}
\newpage
\section{Appendix A.}
\begin{figure}
\begin{center}
\includegraphics[width=0.85\textwidth]{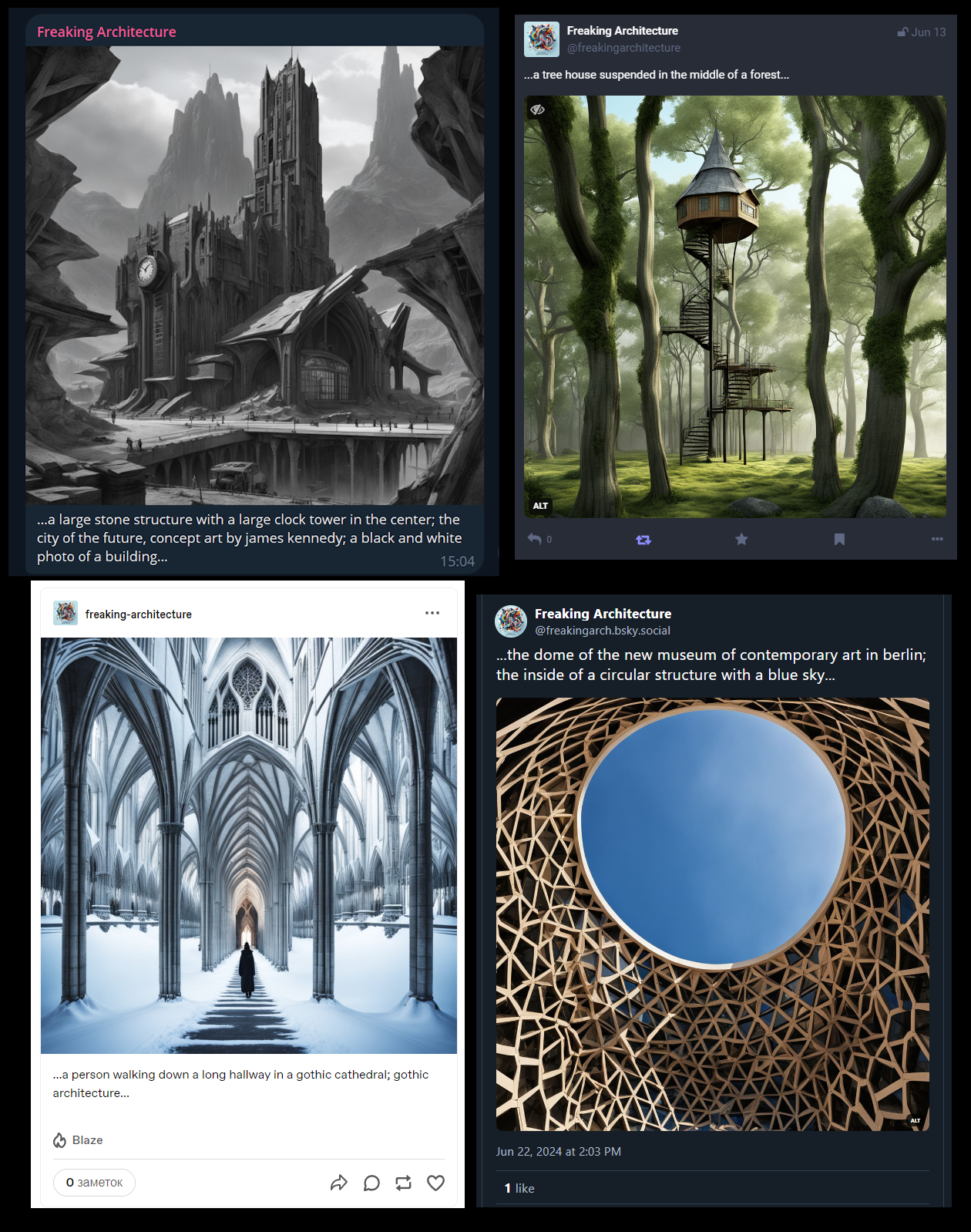}
\end{center}
\caption{Appearance in different publication channels.} \label{pub}
\end{figure}
\end{document}